\def\year{2018}\relax
\documentclass[letterpaper]{article} 
\usepackage{aaai18}  
\usepackage{times}  
\usepackage{helvet}  
\usepackage{courier}  
\usepackage{url}  
\usepackage{graphicx}  
\usepackage{mathtools} 
\usepackage{amsfonts} 
\nocopyright
\begin{document}
%

\title{SPINE: SParse Interpretable Neural Embeddings} 

\author{
Anant Subramanian*, Danish Pruthi*, Harsh Jhamtani*, Taylor Berg-Kirkpatrick, Eduard Hovy\\ 
School of Computer Science \\
Carnegie Mellon University, Pittsburgh, USA\\
\{anant,danish,jharsh,tberg,hovy\}@cmu.edu \thanks{AS, DP and HJ contributed equally to this paper.}
}

\maketitle
\begin{abstract}
Prediction without justification has limited utility.
Much of the success of neural models can be attributed to their ability to learn rich, dense and expressive representations. While these representations capture the underlying complexity and latent trends in the data, they are far from being interpretable.
We propose a novel variant of denoising $k$-sparse autoencoders that generates highly efficient and interpretable distributed word representations (word embeddings), beginning with existing word representations from state-of-the-art methods like GloVe and word2vec.
Through large scale human evaluation, we report that our resulting word embedddings are much more interpretable than the original GloVe and word2vec embeddings. Moreover, our embeddings outperform existing popular word embeddings on a diverse suite of benchmark downstream tasks\footnote{Our code and generated word vectors are publicly available at \url{https://github.com/harsh19/SPINE}}.
\end{abstract}

\section{Introduction}

Distributed representations map words to vectors of real numbers in a continuous space.
These word vectors have been exploited to obtain state-of-the-art results in NLP tasks, such as parsing~\cite{bansal2014tailoring}, named entity recognition~\cite{guo2014revisiting}, and sentiment analysis~\cite{socher2013recursive}. 
However, word vectors have dense representations that humans find difficult to interpret. For instance, we are often clueless as to what a ``high'' value along a given dimension of a vector signifies when compared to a ``low'' value. 
To demonstrate this, we analyze embeddings of few randomly selected words (see Table~\ref{tbl:interpretability}). For these randomly picked words, we examine top participating dimensions (Top participating dimensions are the dimensions that have highest absolute values for that word). For each of these selected top dimensions, we note the words that have the highest absolute values in that dimension. We observe that for embeddings from state-of-the-art word models like GloVe~\cite{glove} and word2vec~\cite{mikolov2013distributed} are not `interpretable', i.e. the top participating words do not form a semantically coherent group. This notion of interpretability --- one that requires each dimension to denote a semantic concept --- resonates with post-hoc interpretability, introduced and discussed in~\cite{lipton2016mythos}.

We argue that this notion of interpretability can help in gaining better understanding of neural representations and models.
Interpretability in a general neural network pipeline would not just help us reason about the outcomes that they predict, but would also provide us cues to make them more efficient and robust.
In various feature norming studies~\cite{garrard2001prototypicality,mcrae2005semantic,vinson2008semantic}, where participants were asked to list the properties of several words and concepts, it was observed that they typically used few \emph{sparse} characteristic properties to describe the words, with limited overlap between different words. For instance, to describe the city of Pittsburgh, one might talk about phenomena typical of the city, like erratic weather and large bridges. It is redundant and inefficient to list negative properties, like the absence of the Statue of Liberty. Thus, sparsity and non-negativity are desirable characteristics of representations, that make them interpretable. Many recent studies back this hypothesis ~\cite{lee1999learning,murphy2012learning,fyshe2014interpretable,fyshe2015compositional,faruqui2015sparse,danish2016discovering}. This raises the following question:

\begin{quotation}
\noindent \emph{
How does one transform word representations to a new space where they are more interpretable?}
\end{quotation}

\def\arraystretch{1}
\begin{table*}[!htb]
\centering
\begin{tabular}{|c|c||c|}\hline
  & \bf{Initial GloVe vectors} & \bf{Initial word2vec vectors} \\ \hline
  \small mathematics & \begin{tabular}{@{}c@{}} \small intelligence, government, foreign, security \\\hline  \small kashmir, algorithms, heat, computational \\\hline \small robes, tito, aviation, backward, dioceses \end{tabular} & \begin{tabular}{@{}c@{}}\small leukemia, enterprises, wingspan, info, booker \\\hline \small ore, greens, badminton, hymns, clay \\\hline \small asylum, intercepted, skater, rb, flats \end{tabular} \\\hline 
  \small remote & \begin{tabular}{@{}c@{}} \small thousands, residents, palestinian, police \\\hline  \small kashmir, algorithms, heat, computational \\\hline \small tamil, guerrilla, spam, rebels, infantry  \end{tabular} & \begin{tabular}{@{}c@{}}\small basilica, sensory, ranger, chapel, memorials\\\hline \small microsoft, sr, malaysia, jan, cruisers\\\hline \small capt, obey, tents, overdose, cognitive, flats \end{tabular} \\\hline 
  \small internet & \begin{tabular}{@{}c@{}} \small thousands, residents, palestinian, police \\\hline  \small intelligence, government, foreign, security \\\hline \small nhl, writer, writers, drama, soccer\end{tabular} & \begin{tabular}{@{}c@{}}\small cardinals, tsar, papal, autobiography, befriends \\\hline \small gases, gov, methane, graph, buttons \\\hline \small longitude, carr, precipitation, snowfall, homer \end{tabular} \\\hline \hline
  & \bf{\texttt{SPOWV} w/ GloVe} \small \cite{faruqui2015sparse} & \bf{\texttt{SPOWV} w/ word2vec} \small \cite{faruqui2015sparse} \\ \hline 
  \small mathematics & \begin{tabular}{@{}c@{}} \small particles, electrons, mathematics, beta, electron\\\hline  \small standardized, wb, broadcasting, abc, motorway \\\hline \small algebra, finite, radcliffe, mathematical, encryption\end{tabular} & \begin{tabular}{@{}c@{}}\small educator, scholar, fluent, mathematician \\\hline \small algebra, instructor, teaches, graduating, graders \\\hline \small batsmen, bowling, universe, mathematician \end{tabular} \\\hline 
  \small remote & \begin{tabular}{@{}c@{}} \small river, showers, mississippi, dakota, format \\\hline  \small haiti, liberia, rwanda, envoy, bhutan \\\hline \small implanted, vaccine, user, registers, lam \end{tabular} & \begin{tabular}{@{}c@{}}\small mountainous, guerrillas, highlands, jungle \\\hline \small pp., md, lightweight, safely, compartment \\\hline \small junk, brewer, brewers, taxation, treaty\end{tabular} \\\hline 
  \small internet & \begin{tabular}{@{}c@{}} \small sandwiches, downloads, mobility, itunes, amazon \\\hline  \small mhz, kw, broadband, licenses, 3g \\\hline \small avid, tom, cpc, chuck ,mori \end{tabular} & \begin{tabular}{@{}c@{}}\small broadcasts, fm, airs, syndicated, broadcast\\\hline \small striker, pace, self, losing, fined \\\hline \small computing, algorithms, nm, binary, silicon \end{tabular} \\\hline \hline
  & \bf{\texttt{SPINE} w/ GloVe} & \bf{\texttt{SPINE} w/ word2vec} \\ \hline 
\small mathematics & \begin{tabular}{@{}c@{}} \small sciences, honorary, faculty, chemistry, bachelor \\\hline  \small university, professor, graduate, degree, bachelor \\\hline \small mathematical, equations, theory, quantum \end{tabular} & \begin{tabular}{@{}c@{}}\small algebra, exam, courses, exams, math \\\hline \small theorem, mathematical, mathematician, equations\\\hline \small doctorate, professor, doctoral, lecturer, sociology \end{tabular} \\\hline 
  \small remote & \begin{tabular}{@{}c@{}} \small territory, region, province, divided, district \\\hline  \small wilderness, ski, camping, mountain, hiking \\\hline \small rugged, mountainous, scenic, wooded, terrain \end{tabular} & \begin{tabular}{@{}c@{}}\small villages, hamlet, villagers, village, huts \\\hline \small mountainous, hilly, impoverished, poorest, populated\\\hline \small button, buttons, click, password, keyboard \end{tabular} \\\hline 
  \small internet & \begin{tabular}{@{}c@{}} \small windows, users, user, software, server \\\hline  \small youtube, myspace, twitter, advertising, ads\\\hline \small wireless, telephone, cellular, cable, broadband \end{tabular} & \begin{tabular}{@{}c@{}}\small hacker, spam, pornographic, cyber, pornography \\\hline \small browser, app, downloads, iphone, download \\\hline \small cellular, subscriber, verizon, broadband, subscribers \end{tabular} \\\hline 
\end{tabular}
\caption{\label{tbl:interpretability} Qualitative evaluation of the generated embeddings. We examine the top participating dimensions for a few randomly sampled words. We then look at the top words from these participating dimensions. Clearly, the embeddings generated by \texttt{SPINE} are significantly more interpretable than both the GloVe and word2vec embeddings, and the Sparse Overcomplete Word Vectors (\texttt{SPOWV})~\cite{faruqui2015sparse}. We also observe that often, the top participating dimensions for a given word are able to cater to different interpretations or senses of the word in question. For instance, for the words `internet' and `remote', we see dimensions that capture different aspects of these words.
}
\end{table*}
\def\arraystretch{1}

\noindent To address the question, in this paper, we make following contributions:
\begin{itemize}
\item
 We employ a denoising $k$-sparse autoencoder to obtain \textbf{SP}arse \textbf{I}nterpretable \textbf{N}eural \textbf{E}mbeddings (\textbf{\texttt{SPINE}}), a transformation of input word embeddings.

We train the autoencoder using a novel learning objective and activation function to attain interpretable and efficient representations. 
\item 
We evaluate \texttt{SPINE} using a large scale, crowdsourced, intrusion detection test, along with a battery of downstream tasks. We note that \texttt{SPINE} is more interpretable and efficient than existing state-of-the-art baseline embeddings.
\end{itemize}

The outline of the rest of the paper is as follows. First, we describe prior work that is closely related to our approach, and highlight the key differences between our approach and existing methods. Next, we provide a mathematical formulation of our proposed method. Thereafter, we describe model training and tuning, and our choice of hyperparameters. Further, we discuss the performance of the embeddings generated by our method on interpretability tests and on various downstream tasks. We conclude by discussing future work.

\section{Related Work}
\label{sec:related}

We first discuss previous efforts to attain interpretability in word representations. Then, we discuss prior work related to $k$-sparse autoencoders.

\subsection*{Interpretability in word embeddings}

Murphy et al.~\shortcite{murphy2012learning} proposed NNSE (Non-Negative Sparse Embeddings) to learn interpretable word embeddings. They proposed methods to learn sparse representations of words using non-negative matrix factorization on the co-occurrence matrix of words. Faruqui et al. \shortcite[a]{faruqui:2015:non-dist} consider linguistically inspired dimensions as a means to induce sparsity and interpretability in word embeddings. However, since their dimensions are binary valued, there is no notion of the \emph{extent} to which a word participates in a particular dimension. Park et al.~\shortcite{park2017rotated} apply rotations to the word vectors to improve the interpretability of the vectors.

Our method is different from these approaches in two ways. Firstly, our method is based on neural models, and is hence more expressive than linear matrix factorization or simple transformations like rotation. Secondly, we allow for different words to participate at varying levels in different dimensions, and these dimensions are discovered naturally during the course of training the network.

Faruqui et al. \shortcite[b]{faruqui2015sparse} have proposed Sparse Overcomplete Word Vectors (\texttt{SPOWV}), that utilizes sparse coding in a dictionary learning setting to obtain sparse, non-negative word embeddings.
Given a set of representations $\mathcal{D} = [\mathbf{X}_1, \mathbf{X}_2, \mathbf{X}_3, \dots, \mathbf{X}_V]^T \in \mathbb{R}^{V \times d}$, where $V$ is the vocabulary size and $d$ is the number of dimensions in the input word embeddings, their approach attempts to represent each input vector $\mathbf{X}_i \in \mathcal{D}$ as a sparse linear combination of basis vectors $\mathbf{a}_j \in \mathbf{A}$. The goal of the Sparse Overcomplete Word Vectors (\texttt{SPOWV}) method is to solve
\begin{align*}
	\operatornamewithlimits{arg\,min}_{\mathbf{D},\mathbf{A}} \big\lVert \mathcal{D} - \mathbf{A} \mathbf{D} \big\rVert_2^2 + \lambda \big\lVert\mathbf{A}\big\rVert_1 + \tau \big\lVert \mathbf{D} \big\rVert_2^2
\end{align*}

\noindent where $\mathbf{D} \in \mathbb{R}^{m \times d}$ is the dictionary of basis vectors, $\mathbf{A}$ is the generated set of sparse output embeddings, and $\lambda$ and $\tau$ are coefficients for the regularization terms. Here, $m$ is the dimensionality of the output embedding space. Sparsity is enforced through the $\ell_1$ penalty imposed on $\mathbf{A}$. The non-negativity constraint is imposed in Faruqui et al.~\shortcite{faruqui2015sparse} during the optimization step, while solving the following equivalent problem.
\begin{align*}
	\operatornamewithlimits{arg\,min}_{\mathbf{D}, \mathbf{A}} \,&\sum_{i=1}^{V} \big\lVert \mathbf{X}_i - \mathbf{a}_i\mathbf{D} \big\rVert_2^2 + \lambda \big\lVert \mathbf{a}_i \big\rVert_1 + \tau \big\lVert \mathbf{D} \big\rVert_2^2 \\
	\text{s.t.} ~\,&\mathbf{D} \,\in \,\mathbb{R}^{m \times d}_{\geq 0} \,~\text{and} \\ 
    &\mathbf{A} \,\in \,\mathbb{R}^{V \times m}_{\geq 0}
\end{align*}

\noindent Here, $\mathbf{A} \in \mathbb{R}^{V \times m}$ is the set of sparse output embeddings. We direct interested readers to \cite{faruqui2015sparse,lee2007efficient} for detailed investigations of this approach.

Their work is similar to ours in spirit and motivation, and is the closest match to our goal of producing interpretable and efficient representations. Thus, we compare our approach with \texttt{SPOWV} in both performance and interpretability tests.

\subsection*{$k$-sparse autoencoders}

A \emph{$k$-sparse} autoencoder \cite{ng2011sparse} is an autoencoder for which, with high probability, at most $k$ hidden units are active for any given input.
Ng \shortcite{ng2011sparse} introduced a mechanism to train $k$-sparse autoencoders. The underlying idea is to achieve an expected activation value for a hidden unit that is equivalent to $k$ completely activated hidden units. The training algorithm does so by augmenting a standard input reconstruction loss with a term that measures the deviation between the observed and the desired mean activation values. However, as \cite{makhzani2013k} note, equating the expected activation values does not necessarily produce exactly (or less than) $k$-sparse representations. 
Our proposed novel objective function and choice of activation function mitigate this issue.


\section{Methodology}
\label{sec:approach}

\def\arraystretch{1.05}
\begin{table}[t]
  \centering
  \small
  \caption{\bf Notation:}
  \begin{tabular}{|lcl|}
    \hline
    $\mathcal{D}$ &: &Input set of representations \\[0.1em]
    $\mathcal{H}$ &: &Set of hidden units in a layer \\[0.1em]
    $Z_h^{(\mathbf{X})}$ &: &Activation value for the hidden unit\\
    && $h$ for the input representation $\mathbf{X}$ \\[0.1em]
    $\rho^*_{h,\mathcal{D}}$ &: &Desired sparsity fraction for unit $h$\\
    && across dataset $\mathcal{D}$\\[0.1em]
    $\rho_{h,\mathcal{D}}$ &: &Observed average activation value \\
    && for unit $h$ across dataset $\mathcal{D}$ \\ \hline
  \end{tabular}
\end{table}
\def\arraystretch{1}

In this section, we describe our neural approach to the task of learning sparse, interpretable neural embeddings (\textbf{\texttt{SPINE}}).\\
Let $\mathcal{D} = [\mathbf{X}_1, \mathbf{X}_2, \mathbf{X}_3, \dots, \mathbf{X}_V]^T \in \mathbb{R}^{V \times d}$ be the set of input word embeddings, where $V$ is the vocabulary size and $d$ is the number of dimensions in the input word embeddings. Our goal is to project these embeddings to a space $\mathbb{R}^m$ such that the $m$-dimensional embeddings in this space are both sparse and non-negative. That is, we wish to find a transformation $\mathbb{R}^{V \times d} \rightarrow \mathbb{R}^{V \times m}$.

\begin{figure*}[ht]
\centering
\includegraphics[width=0.70\textwidth]{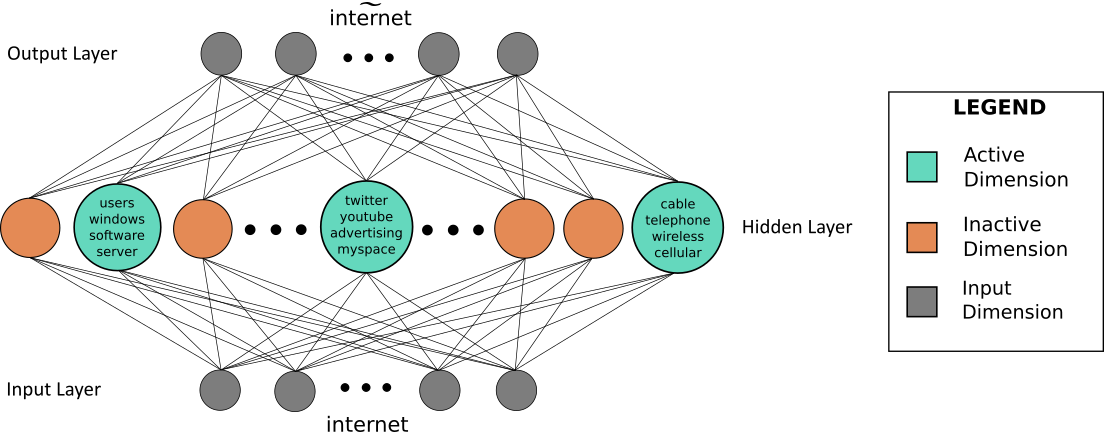}
\caption{Depiction of our $k$-sparse autoencoder for an input word `internet'. Our variant of the $k$-sparse autoencoder attempts to reconstruct the input at its output layer,
with only a few active hidden units (depicted in green). These active units correspond to an interpretable set of dimensions associated with the word `internet'. The rest of the dimensions (depicted in orange) are inactive for this word.}
\label{fig:autoencoder}
\end{figure*}

In contrast to the sparse coding (matrix factorization) approach of \texttt{SPOWV}, we obtain sparse, interpretable embeddings using a neural model. Specifically, we make use of a denoising $k$-sparse autoencoder that is trained to minimize a loss function that concisely captures the required sparsity constraints. The sparse activations corresponding to the given input embeddings are the interpretable vectors generated by our model. Figure~\ref{fig:autoencoder} depicts a $k$-sparse autoencoder that produces sparse and interpretable activations for the input word `internet'.

Let $\widetilde{\mathbf{X}_i}$ be the predicted output for an input embedding $\mathbf{X}_i \in \mathcal{D}$. That is,
\begin{align*}
	Z^{(\mathbf{X}_i)} &= f(\mathbf{X}_i \mathbf{W}_e + \mathbf{b}_e) \\
    \widetilde{\mathbf{X}}_i &= Z^{(\mathbf{X}_i)} \mathbf{W}_o + \mathbf{b}_o
\end{align*}

\noindent where $f$ is an appropriate element-wise activation function, and $\mathbf{W}_e \in \mathbb{R}^{d \times m}$, $\mathbf{W}_o \in \mathbb{R}^{m \times d}$, $\mathbf{b}_e \in \mathbb{R}^{1 \times m}$ and $\mathbf{b}_o \in \mathbb{R}^{1 \times d}$ are model parameters that are learned during optimization. The set $Z = \{Z^{(\mathbf{X}_1)}, Z^{(\mathbf{X}_2)}, \dots, Z^{(\mathbf{X}_m)}\}$ is the set of required sparse embeddings corresponding to each of the input embeddings.

In this setting, given $\mathcal{D}$, our $k$-sparse autoencoder is trained to minimize the following loss function.
\begin{align*}
  \boldmath{L\,(\mathcal{D}) = RL(\mathcal{D}) + \lambda_1 ASL(\mathcal{D}) + \lambda_2 PSL(\mathcal{D})}
\end{align*}

\noindent where $RL(\mathcal{D})$ is the \textbf{reconstruction loss} over the data set, $ASL(\mathcal{D})$ is the \textbf{average sparsity loss} over the data set, and $PSL(\mathcal{D})$ is the \textbf{partial sparsity loss} over the data set. The coefficients $\lambda_1$ and $\lambda_2$ determine the relative importance of the two penalty terms. We define these loss terms below.

\subsubsection{Reconstruction Loss (RL)}
$RL(\mathcal{D})$ is the average loss in reconstructing the input representation from the learned representation. If the reconstructed output for an input vector $\mathbf{X} \in \mathbb{R}^{d}$\, is \,$\widetilde{\mathbf{X}} \in \mathbb{R}^d$, then
    \begin{align*}
      RL\,(\mathcal{D}) = \frac{1}{|\mathcal{D}|} \,\sum_{\mathbf{X} \,\in \,\mathcal{D}} \,\Big\lVert\mathbf{X}-\widetilde{\mathbf{X}}\Big\rVert_{2}^{2}
    \end{align*}

  In the denoising autoencoder setting, we add isotropic Gaussian noise (with mean $\mathbf{0}$, and variance $\sigma^{2}\mathbf{I}$) to enable the autoencoder to learn more robust intermediate representations of the input.

\subsubsection{Average Sparsity Loss (ASL)}
In order to enforce $k$-sparse activations in the hidden layer, \cite{ng2011sparse} describe a modification to the basic autoencoder loss function that penalizes any deviation of the observed average activation value from the desired average activation value of a given hidden unit, over a given data set. We formulate this loss as follows.
    \begin{align*}
      ASL\,(\mathcal{D}) &= \sum_{h \,\in \,\mathcal{H}} \,\Bigl(\max\,\bigl(0, \rho_{h,\mathcal{D}} - \rho^*_{h,\mathcal{D}}\bigr)\Bigr)^2
    \end{align*}

Please note that in addition to the original formulation in~\cite{ng2011sparse}, we also allow for the observed average activation value to be less than the desired average activation value, using a \texttt{max} operator.

\subsubsection{Partial Sparsity Loss (PSL)}
It is possible to obtain an ASL value of $0$ without actually having $k$-sparse representations. For example, to obtain an average activation value of $0.5$ for a given hidden unit across 4 examples, one feasible solution is to have an activation value of $0.5$ for all the four examples (\cite{makhzani2013k} too note this problem).

  To obtain activation values that are truly $k$-sparse, we introduce a novel partial sparsity loss term that penalizes values that are neither close to $0$ nor $1$, pushing them close to either $0$ or $1$. We use the following formulation of PSL to do so.
 \begin{align*}
   PSL\,(\mathcal{D}) &= \frac{1}{|\mathcal{D}|} \,\sum_{\mathbf{X} \,\in \,\mathcal{D}} \sum_{h \,\in \,\mathcal{H}} \Bigl( Z_h^{(\mathbf{X})} \times \bigl(1-Z_h^{(\mathbf{X})}\bigr)\Bigr)
 \end{align*}

 This key addition to the loss term facilitates the generation of sparse embeddings with activations close to 0 and 1.
  
\smallskip
\subsubsection{Choice of activation function}
As motivated earlier, non-negativity in the output embeddings is a useful property in the context of interpretability. This drives us to use activation functions that produce non-negative values for all possible inputs values. The activations produced by Rectified Linear Units (ReLU) and Sigmoid units are necessarily positive, making them promising candidates for our use case. Since we wish to allow for strict sparsity (the possibility of exact $0$ values), we rule out the Sigmoid activation function, due to its asymptotic nature with respect to $0$ activation.

Note that the ReLU activation function produces values in the range $[0,\infty)$, which makes it difficult to argue about the degree of activation of a given hidden unit. Moreover, PSL is not well defined over this range of values. We resolve these issues by using a \textbf{capped ReLU} (\textbf{cap-ReLU}) activation function, that produces activation values in the $[0,1]$ range. Mathematically,
    \begin{align*}
      \text{cap-ReLU}(t) &=
    \begin{dcases}
      \,~0\,, & \text{if }~t\leq 0\\
      \,~t\,, & \text{if }~ 0 < t < 1\\
      \,~1\,, & \text{if }~t\geq 1\\
    \end{dcases}
    \end{align*}


\section{Experimental Setup}
\label{sec:eval}

In this section, we discuss model training, hyperparameter tuning and the baseline embeddings that we compare our method against.

Using the formulation described earlier,
 we train autoencoder models on pre-trained GloVe and word2vec embeddings. The GloVe vectors were trained on 6 billion tokens from a 2014 dump of Wikipedia and Gigaword5, while the word2vec vectors were trained on around 100 billion words from a part of the Google News dataset. Both the GloVe and word2vec embeddings are 300 dimensions long, and we select the 17k most frequently occurring words according to the Leipzig corpus \cite{goldhahn2012building}. We use 15k of these words for training, and use the remaining 2k for hyperparameter tuning.

\subsubsection{Hyperparameter tuning}
\begin{table}[t]
\centering
\begin{tabular}{|c|c|c|c|c|c|}\hline
Vectors & $\rho^{*}$ & $|\mathcal{H}|$ & $\sigma$ & $\lambda_1$ & $\lambda_2$ \\ \hline
GloVe & 0.15 & 1000 & 0.4 & 1 & 1 \\ \hline
word2vec & 0.15 & 1000 & 0.2 & 1 & 1 \\ \hline
\end{tabular}
\caption{Grid-search was performed to select values of the following hyperparamters: Sparsity fraction ($\rho^{*}$), hidden-dimension size ($|\mathcal{H}|$), standard deviation of the additive isotropic zero-mean Gaussian noise ($\sigma$), and the coefficients for the ASL and PSL loss terms ($\lambda_1$ and $\lambda_2$).}
\label{tbl:hyperparameters}
\end{table}
We tune our hyperparameters using the automatic metric to evaluate topic coherence discussed in Lau et al.~\shortcite{lau2014machine}. The metric aims to maximize coherence among different dimensions of the representation, which has been shown by the authors to correlate positively with human evaluation. This is in contrast to~\cite{faruqui2015sparse}, who select hyperparameters that maximize performance on a word similarity task, which does not necessarily correlate with topic coherence. Through experiments with different configurations, we observe that high topic coherence comes at the cost of high reconstruction loss, which manifests itself in the form of poor performance on downstream tasks. To mitigate this issue, we cap the maximum permissible reconstruction loss to a threshold and select the best performing hyperparameter setting within this constraint. The best performing set of hyperparameters are listed in Table~\ref{tbl:hyperparameters}. We observed that a hidden layer of size 1000 units is optimal for our case. Hence, we transform $\mathbf{X} \in \mathbb{R}^{15000 \times 300}$ to $Z \in [0,1]^{15000 \times 1000}$. We also find utility in making the autoencoder denoising, attaining embeddings that are 6\% more sparse at similar reconstruction loss.

Note that through this exercise we evaluate the utility of our additional loss formulation of Partial Sparsity Loss ($PSL$), and we observe that $\lambda_2 = 1$ outperforms $\lambda_2 = 0$ (i.e without the loss) by attaining 22\% more sparsity on GloVe and 67\% more sparsity on word2vec embeddings at comparable reconstruction loss values.  

\subsubsection{Inducing sparsity through $\ell_1$ regularizer}
We experiment with an alternate loss consisting of $\ell_{1}$ regularizer instead of PSL and ASL formulation. In order to achieve similar levels of topic coherence with this formulation, we require very high regularizer coefficient, leading to high reconstruction loss. For a given threshold of reconstruction loss, we observe higher automated topic coherence scores for our proposed ASL \& PSL combination. We attribute the higher score to its ability to force activations to 0 and 1, whereas, $\ell_1$ formulation drives values to 0. Further, $\ell_1$ regularizer is agnostic to the distributions of the sparse values in the embedding.
For example, say we have two-dimensional binary embeddings. Regardless of whether the first dimension in zero for all words, or whether the first dimension is zero for half the words and the second dimension in zero for the rest, $\ell_1$ assigns the same penalty to both. Indeed, we noticed similar trends in our experiments. However, this does not match the hypothesis that every word has a few non-negative characteristics. In contrast, our novel loss formulation does differentiate between the two situations.  

\subsubsection{Baseline embeddings}
We compare the embeddings generated by our model (\texttt{SPINE}) against their corresponding starting embeddings (i.e GloVe and word2vec).
We also compare our word vectors against Sparse Overcomplete Word Vectors (\texttt{SPOWV}) from \cite{faruqui2015sparse}
, which we believe is a more meaningful comparison, as their approach is also tailored to generate sparse, effective, and interpretable embeddings.  In order to perform a fair comparison, we use their method to generate 1000-dimensional output embeddings, the same as ours. All other hyperparameters were used as per the authors' recommendations.

\section{Interpretability}

We evaluate the interpretability of the resulting representations against the ones obtained from baseline models.
We estimate the interpretability of the dimensions in two ways. First, we conduct word intrusion detection tests to quantitatively estimate the interpretability of dimensions in the output embedding space. Second, we qualitatively examine the top participating words from a few randomly sampled dimensions and see if they possess observable similarities.

\subsection{Word Intrusion Detection Test}
\label{sec:intrusiontest}

\begin{figure}[t]
\centering
\includegraphics[width=0.2\textwidth]{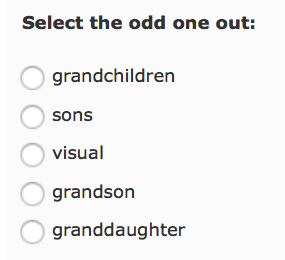}
\caption{A sample intrusion detection question. Here, `visual' is the intruder word.}
\label{fig:amt}
\end{figure}

We adopt the evaluation mechanism from \textit{reading tree leaves} \cite{chang2009reading}, that was first used to interpret probabilistic topic models. Since then, intrusion detection tests have been widely used in estimating interpretability \cite{murphy2012learning}, \cite{fyshe2014interpretable}, \cite{faruqui2015sparse}.

For a given dimension (column) $Z_h$ of the generated embeddings matrix $Z$, we sort the column in the decreasing order of values, and then select the top 4 most active words in that dimension. These four words, if coherent, should be easily identifiable when mixed with a random intruder word. Following the strategies in \cite{murphy2012learning,faruqui2015sparse}, we select a random intruder word that is both present in the bottom half of the dimension $h$ in question, and in the top 10 percentile in at least one other dimension $h' \in \mathcal{H} \setminus \{h\}$. A sample intrusion detection question can be found in Figure~\ref{fig:amt}.

We used the Amazon Mechanical Turk (MTurk) platform to conduct these intrusion detection tests on a large scale. For the two representations \texttt{SPOWV} and \texttt{SPINE}, and for two different initializations (GloVe \& word2vec), we randomly sample 300 of the 1000 dimensions, and pose an intrusion detection test for each of these 300 dimensions. Each Human Intelligence Task (HIT) on MTurk consisted of six such questions. Further, every single HIT was independently solved by three different workers. We also conducted similar intrusion detection tests for each of the original GloVe and word2vec embedding dimensions. Accounting for the various settings, a total of 5400 questions were annotated. 
From the three different annotations we receive for every question, we take the majority vote, and choose that as the answer to that question. In cases where all three annotators mark a different intruder, we randomly select one of the three choices marked by annotators.

\begin{table}[t]
\centering
\small
\begin{tabular}{|c|c|c|}\hline
GloVe & \texttt{SPOWV} & \texttt{SPINE}  \\ 
 (original) &  (w/ GloVe) & (w/ GloVe) \\  \hline
 22.97 & 28.18 & {\bf 68.35}  \\ \hline \hline
 Word2vec & \texttt{SPOWV} & \texttt{SPINE} \\
 (original) &  (w/ word2vec) & (w/ word2vec) \\ \hline
26.08 & 41.75 & {\bf 74.83 } \\ \hline
\end{tabular}
\caption{\label{tbl:amt} Precision scores on the Word Intrusion Detection Test. Higher precision numbers indicate more interpretable dimensions. Clearly, Sparse Interpretable Neural Embeddings (\texttt{SPINE}) outperform the original vectors and Sparse Overcomplete Word Vectors (\texttt{SPOWV}) by a large margin. This is the key result of our work.}
\end{table}

\begin{table}[t]
\centering
\small
\begin{tabular}{|c|c|c|}\hline
GloVe & \texttt{SPOWV} & \texttt{SPINE} \\
 (original) &  (w/ GloVe) & (w/ GloVe) \\  \hline
   76\%, 22\% & 74\%, 21\% & {\bf 83\%}, {\bf 47\%}  \\ \hline \hline
 Word2vec & \texttt{SPOWV} & \texttt{SPINE} \\
 (original) &  (w/ word2vec) & (w/ word2vec) \\ \hline
77\%, 18\% & 79\%, 28\% & {\bf 91\%}, {\bf 48\%}  \\ \hline
\end{tabular}
\caption{\label{tbl:iaa} Inter-annotator agreement across different models, for different starting vectors. In each cell, we list two values: the first one corresponds to the percentage of questions where 2 or more annotators agree, and the second value corresponds to the percentage of questions where all the three annotators agree. Note that in the case of \texttt{SPINE}, nearly half of the times all the annotators agree on a given choice.}
\end{table}

\section{Benchmark Downstream Tasks}
\label{sec:benchmark}

\def\arraystretch{1.25}
\begin{table*}[t]
\centering
\small
\begin{tabular}{|c||@{}c@{}|c|c||@{}c@{}|c|c|}\hline
Task & \begin{tabular}{@{}c@{}} GloVe \\ ~\,(original)~\, \end{tabular} & \begin{tabular}{@{}c@{}} \texttt{SPOWV} \\ (w/ GloVe) \end{tabular} & \begin{tabular}{@{}c@{}} \texttt{SPINE} \\ (w/ GloVe) \end{tabular} & \begin{tabular}{@{}c@{}} \,~Word2vec\,~ \\ (original) \end{tabular} & \begin{tabular}{@{}c@{}} \texttt{SPOWV} \\ (w/ word2vec) \end{tabular} & \begin{tabular}{@{}c@{}} \texttt{SPINE} \\ (w/ word2vec) \end{tabular} \\ \hline
Sentiment Analysis (Accuracy) & 71.37 & 71.83 & {\bf 72.44} & 73.50 & \bf 74.01 & 72.71 \\ \hline
Question Clf. (Accuracy) & 82.80 & {\bf 89.20} & 88.20 & 88.40 & 91.80 & {\bf 92.40} \\ \hline 
Sports News Clf. (Accuracy) & 95.47 & 95.6 & {\bf 96.23} & 92.83 & {\bf 95.6} & 93.96 \\
Religion News Clf. (Accuracy) & 79.35 & 81.72 & {\bf 83.4} & 83.12 & {\bf 84.79} & 82.56  \\
Computers News Clf. (Accuracy) & 71.68 & {\bf 77.86} & 77.47 & 72.71 & {\bf 81.46} & 74.38 \\ \hline
NP Bracketing (Accuracy) & 73.11 & 70.28 & {\bf 74.85} & {\bf 78.13}  & 72.19 & 75.41 \\ \hline
Word Similarity ($\rho$ in \%) & {\bf 66.82} & 66.77 & 64.77 & {\bf 68.42} & 63.73 &  62.79 \\ \hline 
\end{tabular}
\caption{\label{table:downstream} Effectiveness comparison of the generated word embeddings (all accuracies are in \%). We compare the embeddings generated by our \texttt{SPINE} model against the initial GloVe and word2vec vectors, and Sparse Overcomplete Word Vectors (\texttt{SPOWV})
~\cite{faruqui2015sparse} 
on a suite of benchmark downstream tasks.}
\label{tbl:performance}
\end{table*}
\def\arraystretch{1}

To test the quality of the embeddings generated by our model, we use them in the following benchmark downstream classification tasks: sentiment analysis, news classification, noun phrase chunking, and question classification. We also test them using the word similarity task discussed in this section~
(Table \ref{table:downstream}). Like in~\cite{faruqui2015sparse}, we use the average of the word vectors of the words in a sentence as features for text classification tasks. We experiment with SVMs, Logistic Regression and Random forests, which are tuned on the development set. Accuracy is reported on the test set.

\begin{enumerate}

\item  Sentiment Analysis: This task tests the semantic properties of word embeddings. It is a sentence-level binary classification task on the Stanford Sentiment Treebank dataset \cite{socher2013recursive}. We used the provided train, dev. and test splits with only the non-neutral labels, of sizes 8337, 1081 and 2166 sentences respectively.

\item Noun Phrase Bracketing: We evaluate the word vectors on NP bracketing task \cite{lazaridou2013fish}, wherein a noun phrase of 3 words is classified as left bracketed or right bracketed. The NP bracketing dataset contains 2,227 noun phrases split into 10 folds. 
We append the word vectors of three words to get feature representation \cite{faruqui2015sparse}. For words not present in the subset of 17K words we have chosen, we use all zero vectors. We tune on the first fold and report cross-validation accuracy on the remaining nine folds. 

\item Question Classification (TREC): To facilitate research in question answering, \cite{li2006learning} propose a dataset of categorizing questions into six different types, e.g., whether the question is about a location, about a person, or about some numeric information. The TREC dataset comprises of 5,452 labeled training questions, and 500 test questions. By isolating 10\% of the training questions for validation, we use train/validation/test splits of 4906/546/500 questions respectively. 

\item News Classification: Following~\cite{faruqui2015sparse}, we consider three binary
news classification tasks from the 20 Newsgroups dataset\footnote{\url{http://qwone.com/~jason/20Newsgroups/}}. Each task involves categorizing a document according to two related categories (1) Sports: baseball vs. hockey (958/239/796) (2) Computers: IBM vs. Mac (929/239/777) (3) Religion: atheism vs. christian (870/209/717).

\item Word Similarity Task: We use the WS-353 dataset \cite{finkelstein2001placing}, which contains 353 pairs of English words. Each pair of words has been assigned similarity ratings by multiple human annotators. We use the cosine similarity between the embeddings of each pair of words, and report the Spearman's rank correlation coefficient $\rho$ between the human scored list and the predicted similarity list. We consider only those pairs of words where both words are present in the vocabulary. This leads to the removal of $59$ of the $353$ pairs ($17.3\%$).

\end{enumerate}


\section{Results and Discussion}
In this section, we report the results of aforementioned experiments and discuss the implications.

\subsubsection{Interpretability}
Table~\ref{tbl:amt} lists the precision scores of word intrusion detection task for each model with different starting vectors. We observe that our precision scores are notably higher than those of the original vectors, and those of the Sparse Overcomplete Word Vectors. This implies that annotators could select the intruder much more accurately from our dimensions with higher agreement (Table \ref{tbl:iaa}), showing that the resulting representations are highly coherent and interpretable. This forms the key result of our effort to produce more interpretable representations. 
\subsubsection{Performance on downstream tasks}
From the results in Table~\ref{table:downstream}, it is clear that the embeddings generated by our method perform competitively well on all benchmark tasks, and do significantly better on a majority of them.
\subsubsection{Qualitative assessment}
For a few sampled words, we investigate the top words from dimensions where the given word is active. Table~\ref{tbl:interpretability} lists the results of this exercise for three particular words (mathematics, internet and remote), for different models and different starting vectors. From Table~\ref{tbl:interpretability}, we observe that the top dimensions of the word embeddings generated by our model (\texttt{SPINE}) are both coherent, and relevant to the word under examination. Often, our representations are able to capture different interpretations of a given word. For instance, the word `remote' can be used in various settings: remote areas (like remote villages, huts), the electronic remote (like buttons, click), and conditions in remote areas (like poverty). From these examples, we get anecdotal evidence about the higher interpretability achieved by our model on the resulting representations.

\subsubsection{Retroffiting vs joint learning}
We follow a retrofitting approach to obtain sparse and interpretable embeddings. An alternate to this is to add various sparsity and non-negativity inducing regularizers  when training Word2Vec or GloVe. One practical advantage of our retrofitting approach is that it does not need access to a giant corpus. Moreover, our proposed method is agnostic to, and abstracted from, the underlying method used to create the word embeddings, making it more widely applicable.

\subsubsection{Interpretability and downstream performance}

Embeddings from our method outperform original word embeddings for some downstream tasks. This is counter-intuitive, as an increase in interpretability might have come at the expense of downstream performance. We believe that this increased performance is because sparse embeddings directly capture the salient characteristics of concepts, which allow downstream task models to efficiently converge to optimum weight values. For example, if a dimension in the sparse embedding signifies a location name, downstream chunking task models would find it useful.

Further, on choosing 2000 or more hidden dimensions, we found both topic coherence and interpretability to improve, though at a severe cost of performance on downstream tasks. On choosing 500 dimensions, topic coherence and interpretability deteriorated. Theoretically, in the extreme case, one can obtain one-hot vectors by setting the dimension size to be equal to the vocabulary size. These representations would be highly interpretable, but would perform significantly worse on downstream tasks.

\subsubsection{General discussion}
We attribute the success of our method to the expressiveness of a neural autoencoder framework, that facilitates non linear transformations in contrast to existing linear matrix factorization based approaches. We further strengthen the hypothesis that non-negativity and sparsity lead to semantically coherent (interpretable) dimensions.
As per this notion of interpretability, GloVe and word2vec embeddings are highly uninterpretable, whose individual dimensions, by themselves, do not represent any concrete concept or topic. Please note that we do not imply that GloVe and word2vec representations fail to capture the underlying semantic structure. In these representations, similar words are close in the embeddings space, and neighbouring words form a semantically coherent group. In fact, it is due to this characteristic that these representations achieve good scores in word similarity tasks (Table \ref{table:downstream}).

However, we argue that our notion of post-hoc interpretability -- one that requires each dimension to capture a semantic concept -- is a more pragmatic one. In many prediction settings, a softmax layer precedes the class probabilities. Weights from these softmax layers bind to the final layer representation, and large positive and negative weights sway the output class probabilities. In order to explain a prediction, one necessarily has to understand the semantic concepts that each of the dimensions corresponding to these large weights represent. Hence, this notion of post-hoc interpretability is more useful in explaining predictions.


\section{Conclusion}

We have presented a novel mechanism to generate interpretable word embeddings using denoising $k$-sparse autoencoders. Large scale crowd-sourced experiments show that our word embeddings are more interpretable than the embeddings generated by state-of-the-art sparse coding approaches. Also, our embeddings outperform popular baseline representations on a diverse set of downstream tasks.
Our approach uses sub-differentiable loss functions and is trained through back propagation, potentially allowing for seamless integration into neural models, and end-to-end training capabilities. As a part of future work, we are investigating the effect of inducing varying amounts of sparsity at multiple hidden layers in more sophisticated networks, and studying the properties of the resultant sparse activations.

\section{ Acknowledgments}
We thank Dr. Partha Talukdar, Dr. Jason Eisner, Pradeep Dasigi, Dr. Graham Neubig, Dr. Matthew R. Gormley and anonymous AAAI reviewers for their valuable suggestions. The authors also acknowledge Siddharth Dalmia and Mansi Gupta for carefully reviewing the paper draft. DP is supported in part by a Siebel Scholarship and by DARPA Big Mechanism program under ARO contract W911NF-14-1-0436. 

\bibliography{Subramanian-Pruthi-Jhamtani}
\bibliographystyle{aaai}

\end{document}